\pdfoutput=1

\documentclass[11pt]{article}

\usepackage[]{emnlp2021}

\usepackage{times}
\usepackage{latexsym}
\usepackage{todonotes} %
\usepackage{booktabs} 
\usepackage{enumitem}
\usepackage{multirow}
\usepackage{graphicx}
\usepackage{amsmath}
\usepackage{floatrow}
\usepackage{pdflscape}
\usepackage{pifont}
\usepackage{supertabular}
\usepackage{siunitx}
\usepackage{color, colortbl}

\definecolor{cadmiumgreen}{rgb}{0.0, 0.42, 0.24}
\definecolor{carnelian}{rgb}{0.7, 0.11, 0.11}
\newcommand{\cmark}{\ding{51}}%
\newcommand{\xmark}{\ding{55}}%
\usepackage[T1]{fontenc}

\usepackage[utf8]{inputenc}

\usepackage{microtype}

\title{Facebook AI's WMT21 News Translation Task Submission}

\author{
    Chau Tran \enskip
    Shruti Bhosale \enskip
    James Cross \enskip
    Philipp Koehn \enskip
    Sergey Edunov \enskip
    Angela Fan \\
  Facebook AI \\
  \texttt{chau,shru,jcross,pkoehn,edunov,angelafan@fb.com}
}

\begin{document}
\maketitle
\begin{abstract}
We describe Facebook's multilingual model submission to the WMT2021 shared task on news translation. 
We participate in 14 language directions: English to and from Czech, German, Hausa, Icelandic, Japanese, Russian, and Chinese.
To develop systems covering all these directions, we focus on \textit{multilingual} models. 
We utilize data from all available sources --- WMT, large-scale data mining, and in-domain backtranslation --- to create high quality bilingual and multilingual baselines.
Subsequently, we investigate strategies for scaling multilingual model size, such that one system has sufficient capacity for high quality representations of all eight languages. 
Our final submission is an ensemble of dense and sparse Mixture-of-Expert multilingual translation models, followed by finetuning on in-domain news data and noisy channel reranking. 
Compared to previous year's winning submissions, our multilingual system improved the translation quality on all language directions, with an average improvement of 2.0 BLEU. 
In the WMT2021 task, our system ranks first in 10 directions based on automatic evaluation.
\end{abstract}

\section{Introduction}

We participate in the WMT2021 shared task on news translation and submit a multilingual translation system. 
In recent years, multilingual translation has gained significant interest as an alternative to developing separate, specialized systems for different language directions~\cite{firat2016multi,tan2019multilingual,aharoni2019massively,zhang2020improving,tang2020multilingual,arivazhagan2019massively}. 
Multilingual systems have great potential for simplicity and consolidation, making them attractive options for the development and maintenance of commercial translation technologies. 
From a research standpoint, studies of transfer learning between related languages and developing methods that incorporate low-resource languages are strong motivators for grouping languages together in one system~\cite{dabre2019exploiting,fan2021beyond}. 

Despite such motivations, existing multilingual translation systems have been unable to show that the translation quality of multilingual systems surpasses that of bilingual. 
Several works compare to bilingual baselines, but these baselines do not incorporate standard techniques used across the field --- such as backtranslation or dense model scaling. 
Further, multilingual translation systems are often developed on non-standard training datasets and use different evaluation datasets.
These factors make it difficult to assess the performance of multilingual translation, particularly when compared to the most competitive bilingual models.

In this work, our aim is to demonstrate against the winning WMT2020 models and our bilingual WMT2021 systems that multilingual translation models have stronger performance than bilingual ones.
We focus on 14 language directions: English to and from Czech, German, Hausa, Icelandic, Japanese, Russian, and Chinese. 
We create an unconstrained system that utilizes both WMT distributed and publicly available training data, apply large-scale backtranslation, and explore dense and mixture-of-expert architectures. 
We compare the impact of various techniques on bilingual and multilingual systems, demonstrating where multilingual systems have an advantage. 
Our final multilingual submission improves the translation quality on average +2.0 compared to the WMT2020 winning models, and ranks first in 7 directions based on automatic evaluation on the WMT2021 leaderboard. 

\section{Data}

We participate in translation of English to and from Czech (cs), German (de), Hausa (ha), Icelandic (is), Japanese (ja), Russian (ru), and Chinese (zh).
We describe our bitext and monolingual data sources, including additional mined data created for Hausa, and our preprocessing pipeline.

\subsection{Bitext Data}
For all directions, we use all available bitext data from the shared task~\nocite{ziemski2016united}.
For language directions such as English to German or English to Russian, this provides millions of high-quality bitext. 
However, for low to mid resource languages, such as Hausa and Icelandic, we incorporate additional sources of data from freely available online sources such as ccMatrix~\cite{schwenk2019ccmatrix}, ccAligned~\cite{elkishky2020ccaligned}, and OPUS~\cite{tiedemann2012opus}. 
We utilize all available data sources to develop the best quality translation model possible. 

For English-Hausa (and Hausa-English), we also mined extra parallel data from the provided monolingual data. 
We use LaBSE~\cite{feng2020language} to embed Hausa and English sentences into the same embedding space. 
We then use the margin function formulation~\cite{artetxe2019margin} based on $K$-nearest neighbors (KNN) to score and rank pairs of sentences from the two languages. 
Using the mining strategy from~\citet{tran2020cross}, we mined an additional one million pairs of parallel sentences for English-Hausa.

\paragraph{Data Processing.} 
The majority of available bitext represents noisy alignment rather than the output of human translations. 
We apply several steps of preprocessing to filter noisy data. 
First, we apply language identification using \texttt{fasttext}~\cite{joulin2017bag} and retain sentences predicted as the desired language\footnote{Note: for Hausa, the language identification system was unreliable, so we did not utilize it.}.
We then normalize punctuation with \texttt{moses}. 
Subsequently, we removed sentences longer than 250 words and with a source/target length ratio exceeding 3. 

\subsection{Monolingual Data}
Previous work~\citep{ng-etal-2019-facebook} shows that using in-domain monolingual data provides the most quality improvement when used for large-scale backtranslation. 
For high resource languages such as English and German, there are sufficiently large quantities of in-domain data available in Newscrawl, and we do not utilize additional monolingual data.
For the remaining languages, the data available in Newscrawl is not sufficient and we follow the strategy in~\citet{moore2010intelligent,ng-etal-2019-facebook} to examine large quantities of general-domain monolingual data from Commoncrawl\footnote{http://data.statmt.org/cc-100/} and identify a subset that is most similar to the available in-domain news data. 
For each language, we train an n-gram language model~\citep{kenneth2011kenlm} on all available news-domain data (Newscrawl) and a n-gram language model on a similarly sized sample from general-domain data (Commoncrawl). 
For each sentence $s$ in Commoncrawl, we compute word-normalized cross entropy scores $H_{\text{news}}(s)$ and $H_{\text{general}}(s)$ using in-domain language model and general-domain language model respectively. 
We retain sentences that meet the threshold $H_{\text{news}}(s) - H_{\text{general}}(s) > 0.01$.
This selects around 5\% of total number of sentences in the original Commoncrawl. 

\subsection{Vocabulary}
To create our multilingual vocabulary, we first learn a multilingual subword tokenizer on our combined training data across all languages. 
We use SentencePiece~\cite{kudo2018sentencepiece}, which learns subword units from untokenized text.
We train our SPM model with temperature upsampling (with T=5) similar to~\citet{conneau2020unsupervised}, so that low-resource languages are represented. 
Subsequently, we convert the learned SPM units into our final vocabulary. 

\begin{table}[t]
\small
\centering
\begin{tabular}{lrr}
\toprule 
\bf Language & \bf Bitext & \bf Monolingual \\ 
\midrule 
\bf Czech & 185M & 140M \\ 
\bf German & 571M & 237M \\ 
\bf Hausa & 1.7M & 7M  \\ 
\bf Icelandic & 28.2M & 101M \\ 
\bf Japanese & 145.7M & 218M  \\ 
\bf Russian & 297M & 163M  \\ 
\bf Chinese & 166M & 123M  \\ 
\bf English & --- & 430M \\
\bottomrule 
\end{tabular}
\caption{\textbf{Amount of Data per Language}. The bitext data includes data distributed by the WMT Shared Task, the OPUS repository, ccMatrix, ccAligned, and newly mined data for Hausa. The monolingual data includes data distributed by the WMT Shared Task and CC100.}
\label{tab:data_amount}
\end{table}

\section{System Overview}
We describe step-by-step how we created our final multilingual submission for WMT2021. 
We detail our bilingual and multilingual model architectures, as well as how we incorporate strategies such as backtranslation, news-domain finetuning, ensembling, and noisy channel reranking. 

\subsection{Baseline Bilingual Models}
\label{baseline_bilingual_section}

A pre-requisite to creating state-of-the-art multilingual translation systems is establishing strong, competitive bilingual baselines. 
Our goal is to apply the same set of techniques in data augmentation and modeling scaling to both bilingual and multilingual models, and demonstrate multilingual models have stronger translation quality.

To create baseline bilingual systems, we train a separate Transformer model~\cite{vaswani2017attention} for each language direction. 
For every language pair except Hausa, we use the Transformer 12/12 configurations in Table~\ref{tab:dense_multilingual_configs}. For Hausa-English (and English-Hausa), since the amount of bitext data is smaller, we use the Transformer-Base architecture similar to~\citet{vaswani2017attention}.
We train all our models using \texttt{fairseq}~\cite{ott2019fairseq} on 32 Volta 32GB GPUs. We use learning rate of 0.001 with the Adam optimizer, batch size of 768,000 tokens\footnote{6000 tokens per GPU * 32 GPUs * 4 update frequency}, and tune the dropout rate for each language direction independently. For large models

\subsection{Backtranslation}

Backtranslation~\cite{sennrich2015improving} is a widely used technique to improve the quality of  machine translation systems using data augmentation.
To perform backtranslation for a \textit{forward} language direction (e.g. English to German), we use a  system in the backward direction (e.g. German to English), to translate the target German monolingual data into the English source. 
We then use these \textit{backtranslated} synthetic English to German sentence pairs in conjunction with the original parallel data to train an improved forward translation model.

We use all available filtered monolingual data we have for each language (up to 500 million sentences per language) for backtranslation.
Using our baseline bilingual models (described in Section~\ref{baseline_bilingual_section}), we first finetune on in-domain news data (described in Section~\ref{finetuning_section}), and use an ensemble of 3 models with different seeds to generate backtranslation data using beam search. 
For Hausa-English and English-Hausa, we applied a round of iterative backtranslation~\cite{hoang2018iterative, chen2019facebook} as the quality improvement is significant.

\subsection{Data Sharding and Sampling}
\label{data_sharding_section}

Table~\ref{tab:data_amount} displays the amount of data for all languages after postprocessing. 
We divide the data into multiple shards, with each training epoch using one shard. 
We downsample data from both high resource directions and synthetic backtranslated data by dividing them into a greater number of shards than the real bitext data from low resource directions. 
We find that downsampling high resource languages works better than upsampling low resource languages, as upsampling contributes more strongly to overfitting.

\subsection{Model Architectures}
\label{sec:model_arch}
We describe several model architectures that we compared using the final dataset with both bitext and backtranslated data.

\paragraph{Scaling Bilingual Models.}
Based on the baseline architectures described in Section~\ref{baseline_bilingual_section}, we further improve our bilingual models.
The two main improvements are: adding backtranslated data, and adding deeper and wider Transformer configurations to take advantage of the increase in data.

\paragraph{Dense Multilingual Models.}
For the multilingual systems, we train two separate models: \textit{Many to English}, or one system encompassing every language translated into English, and \textit{English to Many}, or one for English into every language.
The challenge of multilingual models is often one of capacity --- given a fixed number of parameters, a model needs to learn representations of numerous languages rather than just one. 
To understand the needed capacity and optimal architectural configuration, we experiment with different Transformer architectures, ranging from 480M parameters to 4.7B parameters (see Table~\ref{tab:dense_multilingual_configs}). 

\begin{table}[t]
\small
\centering
\begin{tabular}{lccc}
\toprule 
\bf & \bf 12/12 & \bf 24/24 & \bf 24/24 Wide 
\\ 
\midrule 
\bf Layers & 12 & 24 & 24\\ 
\bf Emb. Size & 1,024 & 1,024 & 2,048 \\ 
\bf FFN Size & 4,096 & 8,192 & 16,384 \\ 
\bf Attn. Heads & 16 & 16 & 32\\
\midrule 
\bf Total Params. & 480M & 1.2B & 4.7B \\
\bottomrule 
\end{tabular}
\caption{\textbf{Dense Transformer Configurations}.}
\label{tab:dense_multilingual_configs}
\end{table}

\paragraph{Sparsely Gated MoE Multilingual Models.}

In multilingual models, languages necessarily compete for capacity and must balance sharing parameters with specialization for different languages. 
A straightforward way to add capacity to neural architectures is to simply scale the model size in a \textit{dense} manner: increasing the number of layers, the width of the layers, or the size of the hidden dimension. 
However, this has a significant computational cost, as each forward pass activates all parameters --- at the limit, models become incredibly slow to train and produce translations~\cite{fan2021beyond}. 

In this work, we instead focus on \textit{sparse} model scaling, motivated by wanting to increase capacity without a proportional increase in computational cost. 
We train Sparsely Gated Mixture-of-Expert (MoE) models \cite{lepikhin2020gshard} for \textit{English to Many} and \textit{Many to English}. These models aim to strike a balance between allowing high-resource directions to benefit from increased expert model capacity, while also allowing transfer to low-resource directions via shared model capacity.
In each Sparsely Gated MoE layer, each token is routed to the top-k expert FFN blocks based on a learned gating function. Thus, only a subset of all the model's parameters is used per input sequence. 

We use a Transformer architecture with the Feed Forward block in every alternate Transformer layer replaced with a Sparsely Gated Mixture-of-Experts layer with top-2 gating in the encoder and decoder.
As in~\citet{lepikhin2020gshard}, we also add a gate loss term to balance expert assignment across tokens with a gate loss weight of 0.01. We use an expert capacity factor of 2.0. We use a learning rate of 0.001 with the Adam optimizer with 4000 warmup updates and a batch size of 1 Million tokens (MoE model with 64 experts) or 1.5 Million tokens (MoE model with 128 experts).

\begin{table*}[t]
\small
\centering
\begin{tabular}{lccccccc}
\toprule 
    & \bf cs-en & \bf de-en & \bf ha-en & \bf is-en & \bf ja-en & \bf ru-en & \bf zh-en \\
\cmidrule{2-8}
\bf Multilingual Vocab & 28.3 &	\textbf{38.0} &	28.3 &	34.5 &	21.1 &	\textbf{38.0} &	\textbf{30.8} \\
\bf Bilingual Vocab &  \textbf{28.6} &	36.8 &	\textbf{28.4} &	\textbf{35.2} &	\textbf{22.4} &	37.0 &	29.6 \\
\midrule
     & \bf en-cs & \bf en-de & \bf en-ha & \bf en-is & \bf en-ja & \bf en-ru & \bf en-zh \\
\cmidrule{2-8}
\bf Multilingual Vocab & 33.2 &	39.4 &	23.1 &	\textbf{29.4} &	\textbf{26.1} &	25.7 &	42.4  \\
\bf Bilingual Vocab & \textbf{33.7} & \textbf{39.8} & \textbf{23.9} & \textbf{29.4} & \textbf{26.1} &	\textbf{26.0} &	\textbf{43.3}  \\
\bottomrule 
\end{tabular}
\caption{\textbf{Impact of Vocabulary on Bilingual Models}. We compare using a specialized bilingual vocabulary vs. a general multilingual vocabulary and its impact on performance of bilingual systems.}
\label{tab:multilingual_vocab_results}
\end{table*}

\subsection{In-Domain Finetuning}
\label{finetuning_section}
Finetuning with domain-specific data is an effective method of improving translation quality for the desired domain, and thus we curated news-domain data for finetuning.
For directions such as German and Russian, we finetune on evaluation datasets from previous years of WMT.
For Hausa and Icelandic, as no previous data exists, we use mined data and filter to the subset identified as most likely news domain.
Subsequently, we finetune our models on the in-domain data for a maximum of ten epochs, selecting the best model with validation loss on the \texttt{newstest2020} dev set.
For our submission, we use the settings tuned on \texttt{newstest2020} and include \texttt{newstest2021} dev set in the final finetuning. 

\subsection{Checkpoint Averaging}
To combat bias toward recent training data, it is common to average parameters across multiple checkpoints of a model~\cite{vaswani2017attention}. 
We apply this technique to all models and average the last five checkpoints.
To address rapid overfitting during finetuning, we also average the finetuned model with the model after the initial training is complete and select this averaged set of parameters if it performs better on the validation data.

\subsection{Noisy Channel Re-ranking}
We apply noisy channel re-ranking to select the best candidate translations from n-best hypotheses generated with beam search. 
We follow~\citet{yee2019simple,bhosale2020language} and utilize scores from the direct model $P(tgt|src)$, channel model $P(src|tgt)$, and language model $P(tgt)$. To combine these scores for reranking, for every one of our n-best hypotheses, we calculate:
\begin{equation*}
\log P(tgt|src) + \lambda_1 \log P(src|tgt) + \lambda_2 \log P(tgt)
\label{eq:comb}
\end{equation*}
The weights $\lambda_1$ and $\lambda_2$ are determined by tuning
them with a random search over 1000 trials on a validation set and
selecting the weights that give the best performance. In addition, we also tune a length penalty. The search bounds we use for the weights and the length penalty are [0,2].

\paragraph{Language Models.} We trained Transformer-based language models for all languages on the same monolingual data as used for backtranslation. 
The exception is English, where we trained on the CC100 English data and RoBERTa training data~\cite{conneau2020unsupervised,wenzek2019ccnet,liu2019roberta}.
For the high resource languages, the language models have 12 decoder layers and embedding dimension 4096.
For Hausa and Icelandic, we trained smaller language models with 6 decoder layers to prevent overfitting.

\subsection{Post-Processing}
As a final step, we apply post-processing to the translation outputs for Czech, German, Icelandic, Japanese, and Chinese. For Czech, German, and Icelandic, we convert quotation marks to German double-quote style\footnote{https://en.wikipedia.org/wiki/Quotation\_mark\#German}. For Chinese and Japanese, we convert punctuation marks to the language-specific punctuation characters. 

\begin{table*}[t]
\small
\centering
\begin{tabular}{lccccccc|c}
\toprule 
    & \bf cs-en & \bf de-en & \bf ha-en & \bf is-en & \bf ja-en & \bf ru-en & \bf zh-en & \bf Avg \\
\midrule
\bf Bilingual Dense 12/12 & 28.3 & 38.0 & \textbf{28.3} & 34.5 & 21.1 & 38.0 & \textbf{30.8} & \textbf{31.3} \\
\bf Dense 12/12 & 26.9 & 37.5 & \textbf{28.3} & 35.2 & 19.0 & 36.2 & 28.8 & 30.3 \\ 
\bf MoE-64 12/12 & 28.0 & \textbf{38.9} & 27.2 & \textbf{37.3} & 18.5 & \textbf{39.1} & 28.0 & 31.0 \\ 
\cmidrule{2-9}
\bf Dense 24/24 & 28.1 & 37.2 & 26.3 & 35.6 & 20.6 & 35.8 & 28.0 & 30.2 \\ 
\bf MoE-128 24/24 & 28.1 & 36.8 & 23.1 & 36.9 & 18.7 & 36.9 & 29.7 & 29.7 \\ 
\bf Dense 24/24 Wide & \textbf{29} & 37.9 & 24.5 & 36.8 & \textbf{21.2} & 36.9 & 30.4 & 31.0\\
\midrule
\bf Bilingual Dense 12/12, BL-FT & 30.4 & 42.8 & 30.3 & 35.5 & 24.6 & 39.5 & 36.2 & 34.2 \\
\bf Dense 12/12, ML-FT & 30.3 & 42.4 & 32.7 & 37.5 & 23.9 & 39.5 & 34.2 & 34.4 \\ 
\bf MoE-64 12/12, ML-FT & 31.6 & 43.5 & 33.4 & 38.8 & 25.7 & 39.8 & 36.0 & 35.5 \\
\cmidrule{2-9}
\bf Dense 24/24, ML-FT & 31.8 & 43.4 & 36.0 & 38.8 & 25.6 & 40.3 & 36.3 & 36.0 \\ 
\bf MoE-128 24/24, ML-FT & 31.9 & 43.6 & 34.9 & \textbf{39.7} & 26.5 & 40.4 & \textbf{37.2} & 36.3 \\
\bf Dense 24/24 Wide, ML-FT & \textbf{32.1} & \textbf{43.8} & \textbf{36.1} & 39.4 & \textbf{26.7} & \textbf{40.6} & 36.9 & \textbf{36.5}\\
\midrule 
\midrule 
     & \bf en-cs & \bf en-de & \bf en-ha & \bf en-is & \bf en-ja & \bf en-ru & \bf en-zh & \bf Avg \\
\midrule
\bf Bilingual Dense 12/12 & 33.1 & 39.6 & 23.1 & 29.4 & 26.1 & 25.7 & 42.4 & 31.3 \\
\bf Dense 12/12 & 33.7 & 38.6 & 21.4 & 30.5 & 26.6 & 25.3 & 41.1 & 31.0 \\ 
\bf MoE-64 12/12 & 33.5 & 39.7 & 20.4 & 31.5 & 28.0 & 26.4 & 42.5 & 31.7 \\ 
\cmidrule{2-9}
\bf Dense 24/24 & \textbf{34.0} & 39.6 & 21.7 & 31.6 & 27.5 & 26.4 & 42.3 & \textbf{31.9} \\ 
\bf MoE-128 24/24 & 33.0 & \textbf{40.2} & 19.3 & 30.9 & \textbf{28.8} & \textbf{26.6} & \textbf{42.8} & 31.7 \\ 
\bf Dense 24/24 Wide & 33.4 & 39.7 & \textbf{23.4} & \textbf{32.0} & 28.0 & \textbf{26.6} & 42.2 & 32.2 \\
\midrule
\bf Bilingual Dense 12/12, BL-FT & 35.7 & 39.5 & 23.3 & 29.4 & 27.7 & 26.0 & 43.0 & 32.1 \\
\bf Dense 12/12, ML-FT & 35.0 & 39.1 & 22.9 & 30.5 & 26.9 & 25.6 & 41.5 & 31.6 \\ 
\bf MoE-64 12/12, ML-FT & 35.9 & 40.4 & 24.1 & 29.6 & 28.8 & 26.4 & 43.0 & 32.6 \\
\cmidrule{2-9}
\bf Dense 24/24, ML-FT & 35.8 & 40.1 & 24.1 & 31.6 & 28.7 & \textbf{26.8} & 42.5 & 32.8 \\ 
\bf MoE-128 24/24, ML-FT & 36.4 & \textbf{40.8} & \textbf{24.6} & 31.2 & \textbf{29.7} & \textbf{26.8} & \textbf{43.6} & \textbf{33.3} \\
\bf Dense 24/24 Wide, ML-FT & \textbf{36.7} & 40.6 & \textbf{24.6} & \textbf{32} & 29.3 & 26.7 & 43 & \textbf{33.3} \\
\bottomrule 
\end{tabular}
\caption{\textbf{Comparing Dense vs Sparsely Gated MoE Multilingual Models} before and after in-domain fine-tuning. \textit{BL-FT} refers to finetuning a model on bilingual data, while \textit{ML-FT} refers to finetuning a model on multilingual data, see Section~\ref{sect:mlft_blft}.}
\label{tab:moe_vs_dense}
\end{table*}

\begin{table}[t]
\small
\centering
\begin{tabular}{lcc}
\toprule 
    & \bf en-de  \\ 
\midrule 
Bilingual 12/12 & 39.8 \\ 
Bilingual 24/24 & 40.1 \\ 
Bilingual 24/24 Wide & 40.3 \\ 
\midrule 
Bilingual 12/12 + FT & 40.4 \\ 
Bilingual 24/24 + FT & 40.5 \\ 
Bilingual 24/24 Wide + FT & 40.4 \\ 
\bottomrule 
\end{tabular}
\caption{\textbf{Scaling Bilingual Models}.}
\label{tab:scaling_dense}
\end{table}

\section{Experiments and Results}

We conduct experiments to quantify the impact of each of the component in our system. All experiments are evaluated on \texttt{newstest20}~\cite{barrault-etal-2020-findings} using \texttt{SacreBLEU}~\cite{DBLP:journals/corr/abs-1804-08771}.

\subsection{Creating State-of-the-Art Multilingual Translation Models}
\label{sect:mlft_blft}
We investigate the effectiveness of multilinguality in translation. 
Compared to bilingual models, which can dedicate their capacity to specializing in specific source and target languages, multilingual systems must learn to effectively share available capacity across all languages while balancing languages of different resource levels. 
Despite rising research interest, previous WMT submissions have not demonstrated quality improvement of multilingual models over bilingual models. 
We discuss various choices and comparisons that build our state-of-the-art multilingual translation system.
Overall, the best multilingual systems outperform the best bilingual ones in 11 out of 14 directions, with an average improvement of +0.8 BLEU.

\begin{table*}[t]
\small
\centering
\begin{tabular}{lccccccc}
\toprule 
    & \bf cs-en & \bf de-en & \bf ha-en & \bf is-en & \bf ja-en & \bf ru-en & \bf zh-en \\
    \cmidrule{2-8}
\bf Bilingual & 28.3 & 38.0 & 28.3 & 34.5 & 21.1 & 38.0 & 30.8 \\ 
\bf Bilingual, BL-FT & 30.4 & 42.8 & 30.3 &	35.5 &	24.6 &	39.5 &	36.2 \\ 
\cmidrule{2-8}
\bf Multilingual & 29.0 & 37.9 & 24.5 & 36.8 & 21.2 & 36.9 & 30.4 \\ 
\bf Multilingual, BL-FT & 31.8 & 43.3 &	31.9 & 37.0 & 26.5 & \bf 40.6 &	36.8 \\ 
\bf Multilingual, ML-FT & \bf 32.1 & \bf 43.8 &	\bf 36.1 & \bf 39.4 & \bf 26.7 & \bf 40.6 &	\bf 36.9 \\ 
\midrule 
\midrule 
     & \bf en-cs & \bf en-de & \bf en-ha & \bf en-is & \bf en-ja & \bf en-ru & \bf en-zh \\
\cmidrule{2-8}
\bf Bilingual & 33.1 & 39.6 & 23.1 & 29.4 & 26.1 & 25.7 & 42.4 \\ 
\bf Bilingual, BL-FT & 35.7 & 39.5 & 23.3 &	29.4 & 27.7 & 26.0 & \bf 43.0 \\ 
\cmidrule{2-8}
\bf Multilingual & 33.4 & 39.7 & 23.4 &	32.0 & 28.0 & 26.6 & 42.2 \\ 
\bf Multilingual, BL-FT & 36.1 & 40.3 &	24.2 & 30.1 & 28.7 & \bf 27.4 &	\bf 43.0 \\ 
\bf Multilingual, ML-FT & \bf 36.7 & \bf 40.6 &	\bf 24.6 & \bf 32.0 & \bf 29.3 & 26.7 &	\bf 43.0 \\ 
\bottomrule 
\end{tabular}
\caption{\textbf{Impact of Finetuning on Bilingual and Multilingual Models}. \textit{BL-FT} refers to finetuning a multilingual model on bilingual data, while \textit{ML-FT} refers to finetuning a multilingual model on multilingual data.}
\label{tab:finetuning_results}
\end{table*}

\subsubsection{Building a Multilingual Vocabulary.}
Similar to how multilingual systems must share model capacity, multilingual translation models must also share \textit{vocabulary capacity}. 
Instead of training specialized subword units for a specific language (often 32k), multilingual models group all languages together to learn a much smaller vocabulary set than 32k * number of languages. 
We first examine the impact of this multilingual vocabulary, by taking a bilingual system and training it with the multilingual vocabulary. 
This would indicate a performance difference coming not from architecture, but from the vocabulary itself.
Table~\ref{tab:multilingual_vocab_results} indicates that across all directions, using a specialized bilingual vocabulary is usually superior, meaning multilingual systems must bridge the performance gap of a potentially subpar vocabulary. 
However, for some directions such as en-is and en-ja, no difference is observed.

\subsubsection{Comparing Model Architectures.}

\paragraph{Dense Transformer Models.}

Overall, we find that dense multilingual models are fairly competitive with dense bilingual models (see Table~\ref{tab:moe_vs_dense}). 
Importantly, we find multilingual models benefit greatly from additional model capacity.
In Table~\ref{tab:scaling_dense}, we show comparable dense scaling applied to a bilingual model translating from English to German. 
While the multilingual model improves up to 1 BLEU point, the bilingual model only improves +0.3 BLEU, indicating diminishing return and possible overfitting in bilingual models.
Scaling multilingual translation models has stronger potential for performance improvement.

\paragraph{Sparsely Gated Mixture of Expert Models.}

If multilingual models benefit from greater capacity, what is the best way to add that capacity?
In \autoref{tab:moe_vs_dense}, we compare the performance of Dense and MoE multilingual models while keeping the FLOPs per update approximately the same for fair comparison. 
Due to the conditional compute capacity of MoE layers, MoE models have a greater number of total parameters, but a comparable computational cost with the corresponding dense model.

For Many to English and English to Many, the MoE model with 64 experts per MoE layer gives an average boost of +0.7 BLEU on the dev set. 
To compare to scaling dense models, increasing dense model size from 12/12 to 24/24 does not correspond to significant improvement for Many to English. 
However, there is around +1 BLEU improvement in dense scaling on English to Many. 
We also see a slightly decline or no improvement in the performance of MoE models (MoE-64 12/12 vs MoE-128 24/24) when increasing model dimensionality and increasing the number of experts from 64 to 128.
One possible hypothesis is that having 128 experts is largely unnecessary for only 7 languages.
Compared to 64 experts, training convergence per expert is slower as each expert is exposed to fewer tokens during training on an average.

After finetuning on in-domain data, we observe a significant improvement in performance across the board. 
There is a larger improvement from finetuning in MoE models compared to the associated dense baselines.
Furthermore, the MoE model with 128 experts, which previously lagged behind the MoE model with 64 experts, now gives the best results for all but two directions. 
A possible hypothesis is that expert capacity in MoE models can retain specialized direction-specific finetuning better than dense models, where all language directions must share all model capacity while finetuning.

\subsubsection{Effects of In-Domain Finetuning}

\paragraph{Finetuning Improves Multilingual More than Bilingual.} 
Table~\ref{tab:finetuning_results} compares the impact of finetuning across a variety of models. 
Multilingual systems benefit more from in-domain finetuning. 
As a result, the best multilingual system always outperforms the best bilingual system. 

\paragraph{Multilingual Finetuning is better than Bilingual Finetuning.} 
For multilingual models, there are two possible finetuning schemes~\cite{tang2020multilingual}. 
The multilingual model could be finetuned to specialize to the news domain in a multilingual fashion, concatenating the news data for all languages, \textit{or} could be finetuned for each direction separately by training on bilingual news domain data. 
We compare \textit{multilingual in-domain finetuning} with \textit{bilingual in-domain finetuning} in Table~\ref{tab:finetuning_results}. 
We find that multilingual finetuning is almost always better than bilingual finetuning, indicating that it is not necessary to take a multilingual system and specialize it to be bilingual via bilingual finetuning --- a completely multilingual system still has the strongest performance.

\begin{table}[t]
\small
\centering
\begin{tabular}{lccccc}
\toprule 
    & \bf cs-en & \bf de-en & \bf ha-en & \bf is-en & \bf ja-en  \\
\cmidrule{2-6}
\bf Bilingual & 28.9 & 41.5 & 15.9 & 30.3 & 19.7  \\
\bf + BT & 28.3 & 38 & 28.3 &	34.5 & 21.1  \\
\bf $\Delta$ & -0.6 & -3.5 & +12.4 & +4.2 & +1.4 \\
\midrule
     & \bf en-cs & \bf en-de & \bf en-ha & \bf en-is & \bf en-ja \\
\cmidrule{2-6}
\bf Bilingual & 33.1 & 38.7 & 14.7 & 25.8 & 25.4  \\
\bf + BT & 33.2 & 39.4 & 23.1 & 29.4 & 26.1  \\ 
\bf $\Delta$ & +0.1 & +0.7 & +8.4 & +3.6 & +0.7 \\ 
\bottomrule 
\end{tabular}
\caption{\textbf{Impact of Large-scale Backtranslation in Bilingual Systems}. }
\label{tab:backtranslation_bilingual_results}
\end{table}

\begin{table}[t]
\small
\centering
\begin{tabular}{lccccc}
\toprule 
    & \bf cs-en & \bf de-en & \bf ha-en & \bf is-en & \bf ja-en  \\
\cmidrule{2-6}
\bf Multilingual & 27.7 & 37.6 & 16.5 & 34.2 & 20.8 \\
\bf + BT & 27.8 & 37.9 & 25.8 & 35.6 & 20.8 \\ 
\bf $\Delta$ & +0.1 & +0.3 & +9.3 & +1.4 & +0\\ 
\midrule
     & \bf en-cs & \bf en-de & \bf en-ha & \bf en-is & \bf en-ja \\
\cmidrule{2-6}
\bf Multilingual & 33.7 & 39 & 10 & 27 & 26.9\\
\bf + BT & 33.9 & 39.2 & 23.7 & 31.6 & 27.6 \\ 
\bf $\Delta$ & +0.2	& 0.2 & +13.7 & +4.6 & +0.7 \\ 
\bottomrule 
\end{tabular}
\caption{\textbf{Impact of Large-scale Backtranslation in Multilingual Systems}.}
\label{tab:backtranslation_multilingual_results}
\end{table}

\begin{figure}[t]
  \centering
  \includegraphics[width=0.9\linewidth]{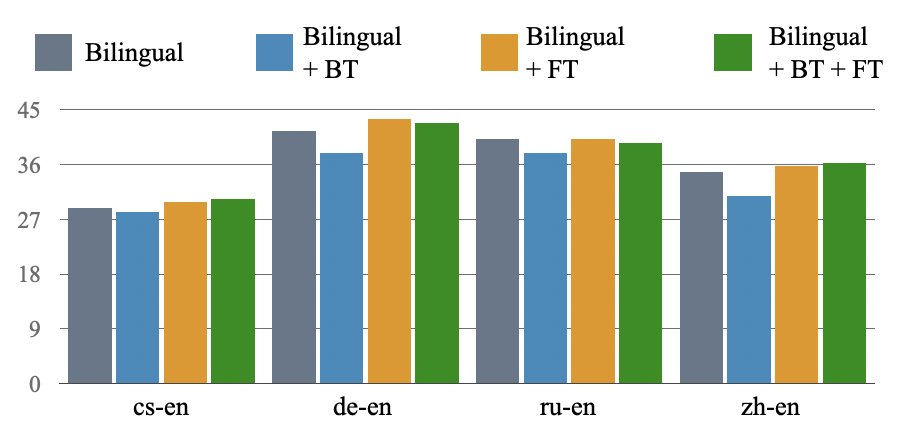}
  \caption{\textbf{Impact of In-Domain Finetuning after Backtranslation} on bilingual models. }
  \label{fig:backtranslation_finetuning_results}
 \end{figure}

\subsubsection{Human Evaluation.}

While a number of studies have been conducted on bilingual models to understand how BLEU correlates with human-perceived quality, few studies have investigated multilingual ones. 
Given a bilingual system and a multilingual system with the same BLEU scores, we want to understand if there is anything intrinsically different in the multilingual system output that would impact human evaluation.

We study two directions: English to German and English to Russian. 
We ask human annotators who are fluent in source and native in  target language to evaluate the translation quality between a bilingual system output and a multilingual system output. 
Both systems have similar BLEU scores, within decimal point difference. 
The translations are generated on the same English source sentence.
We find no statistically significant difference between human evaluations of both systems, indicating that human evaluators have no innate preference for bilingual or multilingual systems.

\begin{table*}[t]
\small
\centering
\begin{tabular}{c|lccccccc|cc}
\toprule 
    \bf MMT & \bf Model & \bf cs-en & \bf de-en & \bf ha-en & \bf is-en & \bf ja-en & \bf ru-en & \bf zh-en & \bf Avg & \bf Incremental $\Delta$ \\
    \midrule 
\textcolor{carnelian}{\xmark} & \bf Bilingual              & 28.9 & 41.5 & 15.9 & 30.3 & 19.7 & 40.2 & 34.8 & 30.2 & --- \\ 
\textcolor{carnelian}{\xmark} & \bf + Backtranslation      & 28.3 &	38.0 & 28.3 & 34.5 & 21.1 &	38.0 & 30.8 & 31.3 & +1.1 \\
\textcolor{carnelian}{\xmark} & \bf + Finetuning           & 30.4 &	42.8 & 30.3 & 35.5 & 24.6 & 39.5 & 36.2	& 34.2 & +2.9 \\
\cmidrule{2-9}
\textcolor{cadmiumgreen}{\cmark} & \bf + Multilingual      & 32.1 &	43.8 & 36.1	& 39.4 & 26.7 & 40.6 & 36.9 & 36.5 & +2.3 \\
\textcolor{cadmiumgreen}{\cmark} & \bf + Ensemble          & 32.3 &	44.5 & 37.2 & 39.9 & 27.2 & 40.9 & 37.8	& 37.1 & +0.6 \\
\textcolor{cadmiumgreen}{\cmark} & \bf + Reranking         & 32.7 &	44.4 & 38.2 & 40.5 & 27.8 & 41.4 & 38.0 & 37.6 & +0.5 \\
\cmidrule{2-11}
\textcolor{carnelian}{\xmark} & \bf WMT20 Winner	       & 29.9 & 43.8 & ---  & ---  & 26.6 & 39.2 & 36.9 &   &   \\
& \bf $\Delta$ over WMT20 & +2.8 & +0.6 & ---  & ---  & +1.2 & +2.2 & +1.1 &   &   \\
\toprule 
\toprule 
    \bf MMT & \bf Model & \bf en-cs & \bf en-de & \bf en-ha & \bf en-is & \bf en-ja & \bf en-ru & \bf en-zh & \bf Avg & \bf Incremental $\Delta$ \\
    \midrule 
\textcolor{carnelian}{\xmark} & \bf Bilingual  & 33.1 &	38.7 &	14.7 &	25.8 &	25.4 &	25.8 &	40.0  &	29.1 & ---            \\ 
\textcolor{carnelian}{\xmark} & \bf + Backtranslation & 33.1 &	39.6 &	23.1 &	29.4 &	26.1 &	25.7 &	42.4 &	31.3 &	+2.3     \\
\textcolor{carnelian}{\xmark} & \bf + Finetuning & 35.7 &	39.5 &	23.3 &	29.4 &	27.7 &	26.0 &	43.0 &	32.1 &	+0.7         \\
\cmidrule{2-9}
\textcolor{cadmiumgreen}{\cmark} & \bf + Multilingual & 36.4 &	40.8 &	24.6 &	31.2 &	29.7 &	26.8 &	43.6 &	33.3 &	+1.2     \\
\textcolor{cadmiumgreen}{\cmark} & \bf + Ensemble &   36.8 &	41.1 &	25.0 &	32.5 &	29.7 &	26.9 &	43.6 &	33.7 &	+0.4    \\
\textcolor{cadmiumgreen}{\cmark} & \bf + Reranking &  37.2 &	41.1 &	25.5 &	32.8 &	29.7 &	27.4 &	43.6 &	33.9 &	+0.2      \\
\textcolor{cadmiumgreen}{\cmark} & \bf + Postprocessing &   39.8 &	42.6 &	25.5 &	34.5 &	29.8 &	28.8 &	48.2 &	35.6 &	+1.7 \\ 
\cmidrule{2-11}
\textcolor{carnelian}{\xmark} & \bf WMT20 Winner & 36.8 &	38.8 & --- & --- &	28.4 &	25.5 &	47.3		       \\
& \bf $\Delta$ over WMT20 &  +3.0 &	+3.8 & --- & --- &	+1.4 &	+3.3 &	+0.9 \\
\bottomrule 
\end{tabular}
\caption{\textbf{Full Results of Submitted Models}. Starting with a bilingual baseline, we depict the incremental gain of different techniques across language pairs. Our final submission is a multilingual ensemble with noisy channel reranking, trained on all available data including backtranslation. On all language pairs, we observe improvement compared to the previous WMT20 winning models. The column \textit{MMT} denotes if the model is multilingual. Note Hausa and Icelandic were not present in WMT20.}
\label{tab:full_results}
\end{table*}

\subsection{Impact of Large-scale Backtranslation}
\label{backtranslation_results_section}

Large-scale backtranslation has contributed to improvements in performance in machine translation models~\cite{edunov2018understanding}, even when measured in human evaluation studies~\cite{edunov2019evaluation, bogoychev2019domain} --- it is a component integrated into most modern translation systems.
However, backtranslation also has downsides. 
Research has indicated that systems trained with large scale backtranslation data tend to overfit to the synthetically generated source sentences, producing lower quality translations when translating original source sentences~\cite{marie2020tagged}.
Further, backtranslation is fundamentally a form of data augmentation, which could have increasingly marginal effect when large-scale mined bitext is directly incorporated into training datasets. 
Beyond mining, multilingual translation can also be seen as an inherent form of data augmentation, as language directions can benefit from the training data of other directions. 
Thus, we analyze further in this section the continued importance of backtranslation, even in multilingual systems.

\paragraph{Backtranslation in Bilingual Systems.}

First, we investigate if backtranslated data is still helpful, even after we augment the training dataset with mined and publicly available training data, beyond what is distributed in the WMT Shared Task.
Our results in Table~\ref{tab:backtranslation_bilingual_results} show that backtranslation is helpful for 10 out of 14 directions, especially for low resource directions such as ha-en and is-en. 
However, for high resource directions such as de-en, ru-en, zh-en, bilingual systems trained with backtranslation had slightly lower validation BLEU compared to those trained without backtranslation.

\paragraph{Finetuning Corrects Overfitting to Translationese}

We further investigate the anomaly that high-resource directions can suffer from adding backtranslated data. 
Figure~\ref{fig:backtranslation_finetuning_results} shows that the minor BLEU degradation from adding backtranslation mostly disappears after applying in-domain finetuning. 
For zh-en and cs-en after in-domain finetuning, the system trained with backtranslation has stronger performance (+0.4 BLEU) compared to the system trained without backtranslation.
Previous studies of this effect have indicated that backtranslation produces \textit{translationese}, which has distinct qualities compared to original training data~\cite{marie2020tagged, zhang2019effect, graham2020translationese}.
We hypothesize that in-domain finetuning, which trains the model on non-backtranslated data, can have a \textit{corrective} effect that counteracts overfitting on translationese.

\paragraph{Backtranslation in Multilingual Systems.}

Table~\ref{tab:backtranslation_multilingual_results} summarizes the performance improvement from adding backtranslation to multilingual models in an ablation study. 
Overall, despite creating a fully unconstrained system with substantially greater training data and leveraging the data sharing potential of multilingual translation, we find that backtranslation still improves the performance. 
We believe this is influenced by the fact that backtranslation fully utilizes available monolingual data.
While data mining techniques can identify potentially parallel sentences, it is naturally limited to identifying only a subset of the full monolingual data the algorithms utilize to mine.

\subsection{Ablation on Components of Final Submission}

Finally, we end by analyzing each aspect in our final submission and the cumulative effect.
The effect of each component is shown in Table~\ref{tab:full_results}. 

\paragraph{Bilingual Baselines.} 
We find that our bilingual baselines have high BLEU scores, particularly for ru-en where our bilingual baseline is already stronger than the WMT20 winner. 
Overall, we observe that only en-ha and ha-en are significantly lower than 20 BLEU, indicating that curating a large amount of high quality bitext data is likely the most important basis of a strong  system.

\paragraph{Backtranslation.} 
Subsequently, we add backtranslated data.
We observe that ha, is, and ja in particular observe large improvements in BLEU after adding backtranslated data, while other directions can actually slightly decrease in quality as a possible effect of translationese.

\paragraph{In-Domain Finetuning.}
We next evaluate the impact of in-domain finetuning and find an almost 3 BLEU improvement across directions for translation into English and 0.7 BLEU improvement for translation out of English.
Across all language directions, finetuning is almost universally helpful. 

\paragraph{Multilingual.} 
Compared to bilingual models, multilingual models have stronger performance in every direction. Multilingual models benefit much more from scaling model size, as our largest architecture (MoE-128 24/24) has the best performance.

\paragraph{Ensembling.}
The effect of ensembling on average is fairly minor, but specific directions can see large improvements (such as +1 BLEU on zh-en).

\paragraph{Reranking.} 
We then apply noisy channel reranking to the outputs of our final system. 
It is helpful across almost all directions, but does not have a huge effect on BLEU. 
On average, performance improves around 0.3 to 0.5 BLEU.

\paragraph{Postprocessing.}
Finally, we observe that postprocessing translated output to use standardized punctuation in each language is very important for BLEU scores when translating out of English.
For example, Chinese in particular has a number of specific periods and double width punctuation characters, and properly using these produces almost +5 BLEU. 
However, we note that these techniques likely only improve BLEU score, and the effect on human evaluation is not well understood.

\section{Conclusion}
In this paper, we describe Facebook's multilingual model submission to the WMT2021 shared task on news translation. 
We employed techniques such as large scale backtranslation, bitext mining, large scale dense and sparse multilingual models, in-domain finetuning, ensembling, and noisy channel reranking. 
We provide extensive experiment results to quantify the impact of each technique, as well as how well they cumulatively stack to produce the final system. 
Our results demonstrate that multilingual translation can achieve state-of-the-art performance on both low resource and high resource languages, beating our strong bilingual baselines and previous years' winning submissions.

\section*{Acknowledgements}
We’d like to thank Michael Auli and Halil Akin for helping getting this project started, help and advice along the way. We'd like to thank Holger Schwenk and Vishrav Chaudhary for their help in getting training data. 

\bibliography{anthology,custom}
\bibliographystyle{acl_natbib}

\end{document}